\documentclass[pdftex,publish,11pt]{interact}
\usepackage[numbers,sort&compress]{natbib}
\bibpunct[, ]{[}{]}{,}{n}{,}{,}
\renewcommand\bibfont{\fontsize{10}{12}\selectfont}

\makeatletter
\def\NAT@def@citea{\def\@citea{\NAT@separator}}
\makeatother
\usepackage{multirow}
\usepackage{url}
\theoremstyle{plain}

\theoremstyle{definition}

\theoremstyle{remark}

\begin{document}

\articletype{FULL PAPER}

\title{Shepherding UAV Swarm with Action Prediction Based on Movement Constraints}

\author{
\name{Yusuke Tsunoda\textsuperscript{a}\thanks{CONTACT Yusuke Tsunoda Email: tsunoda@eng.u-hyogo.ac.jp}, Yusuke Goto\textsuperscript{a},and Takao Sato\textsuperscript{a}}
\affil{\textsuperscript{a}Graduate School of Engineering, University of Hyogo, Hyogo, Japan;}
}

\maketitle

\begin{abstract}
In this study, we propose a new sheepdog-inspired control method for a swarm of small unmanned aerial vehicles (UAVs), which predicts the swarm behavior while explicitly accounting for the motion constraints of real robots.
Sheepdog-inspired guidance control refers to a framework in which a small number of navigator agents (sheepdog agents) indirectly drive a large number of autonomous agents (a flock of sheep agents) so as to steer the group toward a target position.
In conventional studies on sheepdog-inspired guidance, both types of agents have typically been modeled as point masses, and the guidance law for the navigator agents has been designed using simple interaction vectors based on the instantaneous relative positions between the agents.
However, when implementing such methods on real robots such as drones, it is necessary to consider each agent's motion constraints, including upper bounds on velocity and acceleration.
Moreover, we argue that guidance can be made more efficient by predicting the future behavior of the autonomous swarm that is observable to the navigator agents.
To this end, we propose a three-dimensional guidance control law based on behavior prediction of autonomous agents under motion constraints, inspired by the Dynamic Window Approach (DWA).
At each control cycle, the navigator agent generates a set of feasible motion candidates that satisfy its motion constraints, and predicts the short-horizon swarm evolution using an internal model of the autonomous agents maintained within the navigator agent.
The motion candidates are then evaluated according to criteria such as the progress velocity toward the target, the positioning strategy with respect to the swarm, and safety margins, and the optimal motion is selected to achieve safe and efficient guidance.
Numerical simulation results demonstrate the effectiveness of the proposed guidance control law.
\end{abstract}

\begin{keywords}
Swarm robotics, Unmanned Aerial Vehicle, Shepherding, Predictive control, Motion constraint
\end{keywords}

Unmanned aerial vehicles (UAVs), particularly multicopter-type UAVs (drones), have attracted increasing attention in recent years due to rapid advances in miniaturization and cost reduction, and swarm control of drones has been actively studied as an advanced operational paradigm.
In swarm guidance methods, an autonomous distributed control system is often adopted, where each agent determines its actions based on locally available observations, and the desired collective guidance behavior emerges as a result.
Since this scheme does not require each robot to communicate with all other robots as a prerequisite, it offers excellent scalability and enables the construction of systems that are inherently robust.
However, because each agent must decide its behavior using only limited information, the design of the underlying ``local interaction rules'' becomes a fundamental yet challenging issue.
As a design principle, bio-inspired approaches that draw on collective animal behaviors have been widely explored.
Among them, sheepdog-inspired swarm control has received considerable attention in recent years.
Sheepdog-inspired swarm control is motivated by ``shepherding,'' in which sheepdogs drive a flock of sheep, and it is a system in which a small number of \emph{navigator agents} (corresponding to sheepdogs) approach a large \emph{swarm of autonomous agents} (corresponding to sheep) and induce their evasive behavior, thereby steering the swarm toward a target location~\cite{shepherding_survey}.
In particular, only the navigator agents have access to information about the target, and they exploit their superior mobility to actively approach the autonomous agents, whereas the autonomous agents act solely based on their own local observations.
Consequently, communication capabilities can ideally be concentrated in the navigator agents, which is expected to reduce communication load and the overall system cost when implemented in real drone swarm control.

\begin{figure}[t]
    \centering
    \includegraphics[width=\linewidth]{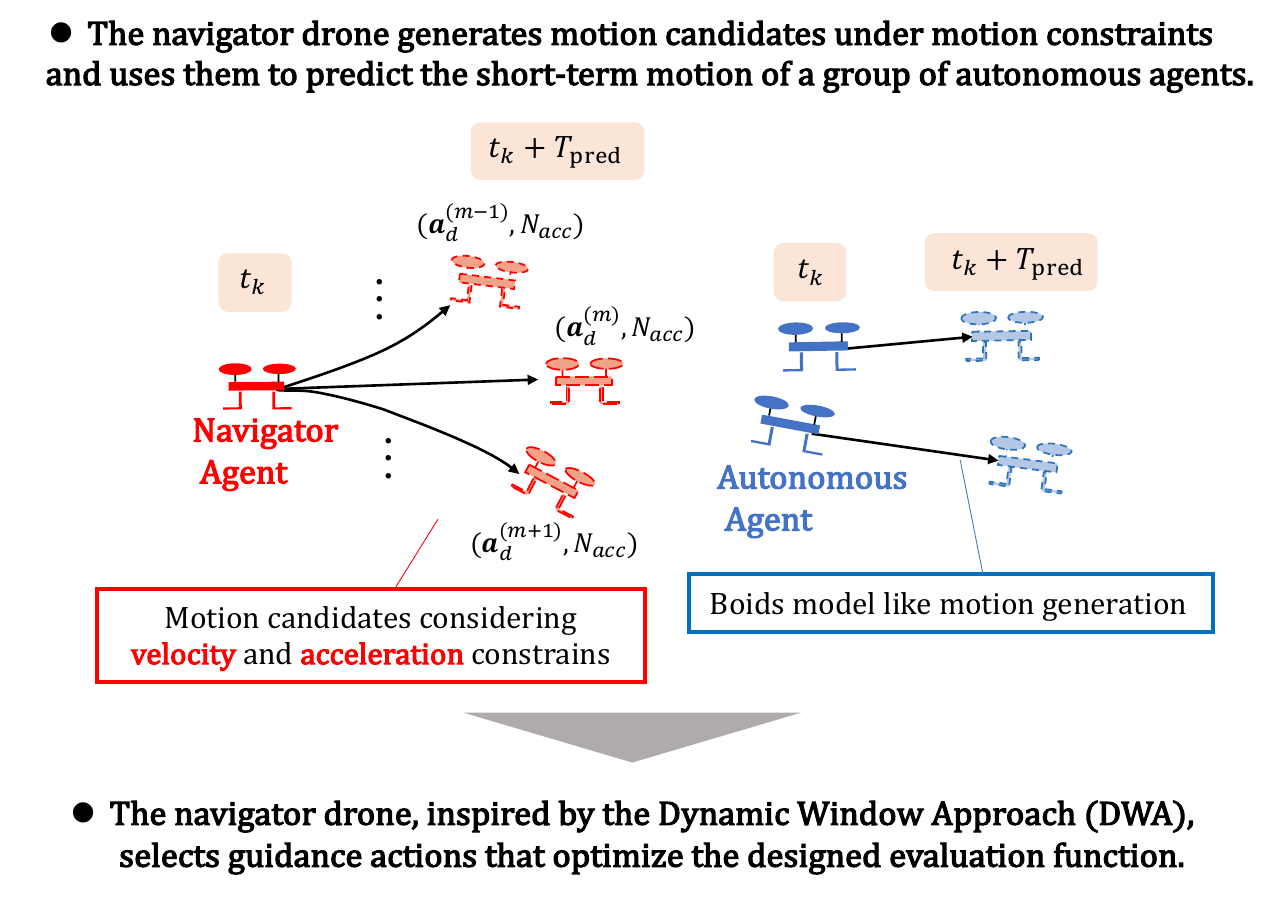}
    \caption{Conceptual diagram of the proposed approach. The navigator agent generates candidate motions that satisfy its motion constraints, predicts the swarm behavior under each candidate motion using an internal model of the autonomous agents, evaluates the candidates based on criteria such as progress toward the goal and safety margins, and selects the optimal motion to achieve safe and efficient guidance.}
    \label{fig:approach}
\end{figure}

A substantial body of research on sheepdog-inspired swarm control already exists.
Most studies model the guided autonomous swarm using flocking models such as the boids model and then propose guidance laws for the navigator agents that can steer the swarm efficiently.
A wide variety of guidance strategies have been reported, including the centroid-tracking approach~\cite{tsunoda2018analysis,Yusuke_TSUNODA2019}, dynamic target switching~\cite{strom}, farthest-agent targeting control~\cite{tsunoda2018analysis}, and the outmost pushing strategy~\cite{zhang2024distributed}.
However, many existing studies assume that the navigator agents can acquire the surrounding environment and the positions of autonomous agents without limitation, and/or that the robots can instantaneously generate ideal velocities and accelerations.
As a result, system models that explicitly incorporate the motion constraints and performance limits of real robots have not been sufficiently established.
In addition, guidance control is often designed as a simple interaction model based only on the instantaneous relative positions among agents at each time step.
While such simplification can enable effective guidance for a particular assumed swarm model with a simple strategy, it may leave little room to design guidance controllers that generalize to more diverse models of autonomous swarms.

In this study, we revisit the conventional shepherding system model and propose a new prediction-based shepherding control method for a swarm of small UAVs that explicitly leverages the motion constraints of the autonomous agents.
With implementation on real drone swarms in mind, we impose upper bounds on the magnitudes of velocity and acceleration for each agent as shown in Fig.~\ref{fig:approach}.
Building on this constrained model, and inspired by the Dynamic Window Approach (DWA), we present a guidance strategy in which a navigator agent predicts the possible behaviors of the autonomous agents in response to its own candidate actions and then selects the action that minimizes (or maximizes) a predefined evaluation function.
A related line of work on shepherding with model predictive control is reported in~\cite{2022_ogura}; however, that study focuses on a different problem setting, namely the presence of heterogeneous agents embedded in the autonomous swarm that do not exhibit avoidance behavior toward the sheepdog agents.
Numerical simulation results demonstrate the effectiveness of the proposed guidance system.

The remainder of this paper is organized as follows.
Section~\ref{sec:problem} describes the problem formulation of the shepherding task considered in this study.
Section~\ref{sec:sheep} presents the behavioral model of the autonomous swarm to be guided.
Section~\ref{sec:shepherd} explains the guidance strategy for the navigator agent, which is based on predicting the autonomous swarm behavior under motion constraints, inspired by the Dynamic Window Approach (DWA).
Section~\ref{sec:sim} reports numerical simulation results of the proposed system and provides discussions.
Finally, Section~\ref{sec:conclusion} concludes the paper.

Throughout this paper, we use the following notation.
Let $\mathbb{R}$ denote the set of real numbers, $\mathbb{R}_+$ the set of positive real numbers, and $\mathbb{N}$ the set of natural numbers.
Let $\mathbb{R}^n$ denote the set of $n$-dimensional real-valued vectors, and let $\|\cdot\|$ denote the $L^2$-norm.

\section{Problem Formulation}
\label{sec:problem}
In sheepdog-inspired guidance, cooperation among multiple navigator agents is an important research topic.
As a fundamental step, however, this study focuses on a sheepdog system consisting of a single navigator agent and $N\in\mathbb{N}$ autonomous agents.
Each agent is modeled as a second-order integrator subject to velocity and acceleration constraints.
Let $\Delta t\in\mathbb{R}_+\,[\mathrm{s}]$ be the sampling period, let $k=0,1,2,\ldots$ be the discrete-time index, and let $t_k = k\Delta t\,[\mathrm{s}]$ be the discrete time.
When an input acceleration $\bm{a}(t_k)\,[\mathrm{m/s^2}]$ is applied at each step, the state is updated according to the following difference equations.
The equations of motion for the autonomous agents and the navigator agent are given in~(\ref{eq:sheep}) and~(\ref{eq:dog}), respectively.

\begin{equation}
  \label{eq:sheep}
  \begin{aligned}
    \begin{cases}
    \bm{p}_{s,i}(t_{k+1}) = \bm{p}_{s,i}(t_k) + \bm{v}_{s,i}(t_k)\Delta t\\
    \bm{v}_{s,i}(t_{k+1}) = \bm{v}_{s,i}(t_k) + \bm{a}_{s,i}(t_k)\Delta t\\
    \|\bm{v}_{s,i\_\rm{hor}}(t_k)\| \leq v_{s,\mathrm{max\_hor}},\\\qquad
    -v_{s,\mathrm{max\_down}} \leq v_{s,i\_\rm{ver}}(t_k) \leq v_{s,\mathrm{max\_up}}\\
    \|\bm{a}_{s,i\_\rm{hor}}(t_k)\| \leq a_{s,\mathrm{max\_hor}},\\\qquad
    -a_{s,\mathrm{max\_down}} \leq a_{s,i\_\rm{ver}}(t_k) \leq a_{s,\mathrm{max\_up}}
    \end{cases}\\
    \qquad i = 1,2,\ldots,N,
  \end{aligned}
\end{equation}

\begin{equation}
    \label{eq:dog}
    \begin{aligned}
    \begin{cases}
        \bm{p}_d(t_{k+1}) = \bm{p}_d(t_k) + \bm{v}_d(t_k)\Delta t,\\
        \bm{v}_d(t_{k+1}) = \bm{v}_d(t_k) + \bm{a}_d(t_k)\Delta t,\\
        \|\bm{v}_{d\_\rm{hor}}(t_k)\| \leq v_{d,\mathrm{max\_hor}},\\\qquad
        -v_{d,\mathrm{max\_down}} \leq v_{d\_\rm{ver}}(t_k) \leq v_{d,\mathrm{max\_up}},\\
        \|\bm{a}_{d\_\rm{hor}}(t_k)\| \leq a_{d,\mathrm{max\_hor}},\\\qquad
        -a_{d,\mathrm{max\_down}} \leq a_{d\_\rm{ver}}(t_k) \leq a_{d,\mathrm{max\_up},}
    \end{cases}
    \end{aligned}
\end{equation}
where $\bm{p}\in\mathbb{R}^3\,[\mathrm{m}]$ denotes the position vector, $\bm{v}\in\mathbb{R}^3\,[\mathrm{m/s}]$ the velocity vector, and $\bm{a}\in\mathbb{R}^3\,[\mathrm{m/s^2}]$ the acceleration vector.
The subscript $s,i$ refers to the $i$-th autonomous agent, while the subscript $d$ refers to the navigator agent.
In addition, the superscripts $\mathrm{hor}$ and $\mathrm{ver}$ indicate the horizontal and vertical components, respectively.
To allow different maximum values for ascending and descending motions, we impose the vertical velocity constraint in the form
$-v_{\ast,\mathrm{max\_down}} \le v_{\ast,\mathrm{ver}}(t_k) \le v_{\ast,\mathrm{max\_up}}$
for $\ast\in\{s,d\}$.
Similarly, the vertical acceleration is constrained as
$-a_{\ast,\mathrm{max\_down}} \le a_{\ast,\mathrm{ver}}(t_k) \le a_{\ast,\mathrm{max\_up}}$.

The objective of this study is to guide the multi-agent system governed by the above equations of motion by designing the autonomous-agent inputs $\{\bm{a}_{s,i}(t_k)\}_{i=1}^{N}$ and the navigator-agent input $\bm{a}_{d}(t_k)$ such that the target-reaching condition~(\ref{eq:goal}) and the collision-avoidance constraint~(\ref{eq:collision_avoid}) shown below are satisfied simultaneously.
\begin{equation}
    \label{eq:goal}
    \lim_{k\to\infty}\|\bm{p}_{s,i}(t_k) - \bm{p}_{\mathrm{goal}}\| \to \varepsilon, \quad i = 1, 2, \ldots, N,
\end{equation}

Here, $\bm{p}_{\mathrm{goal}}\in\mathbb{R}^3\,[\mathrm{m}]$ denotes the position vector of the goal location, and $\varepsilon\in\mathbb{R}_+$ is a sufficiently small positive constant.
From the perspective of implementation on real drones, each agent is approximated as a rigid sphere with an identical radius $r_{\mathrm{indiv}}\in\mathbb{R}_+\,[\mathrm{m}]$.
We require that the occupied regions of any two agents do not overlap at any time.
That is, we impose the collision-avoidance constraint given by

\begin{equation}
\label{eq:collision_avoid}
\begin{aligned}
&\|\bm{p}_{s,i}(t_k)-\bm{p}_{s,j}(t_k)\| \ge 2r_{\mathrm{indiv}},\\
&\qquad \forall k\ge0,\ \forall i,j\in\{1,\ldots,N\},\ i\neq j,\\
&\|\bm{p}_{d}(t_k)-\bm{p}_{s,i}(t_k)\| \ge 2r_{\mathrm{indiv}},\\
&\qquad \forall k\ge0,\ \forall i\in\{1,\ldots,N\}.
\end{aligned}
\end{equation}

Regarding the sensing mechanism and the information available to each agent, we assume a common observation radius of $R_{s}\in\mathbb{R}_+\,[\mathrm{m}]$ for the autonomous agents and an observation radius of $R_{d}\in\mathbb{R}_+\,[\mathrm{m}]$ for the navigator agent.
Each agent determines its action by referencing the states of agents within its sensing range while distinguishing the agent type (autonomous agent or navigator).
In addition to this local information, the navigator agent is assumed to have access to the position of the goal location.
Equation~(\ref{eq:obs}) defines the observation model for each agent and the construction of the input $\bm{a}(t_k)$.

\begin{figure}[t]
    \centering
    \includegraphics[width=0.7\linewidth]{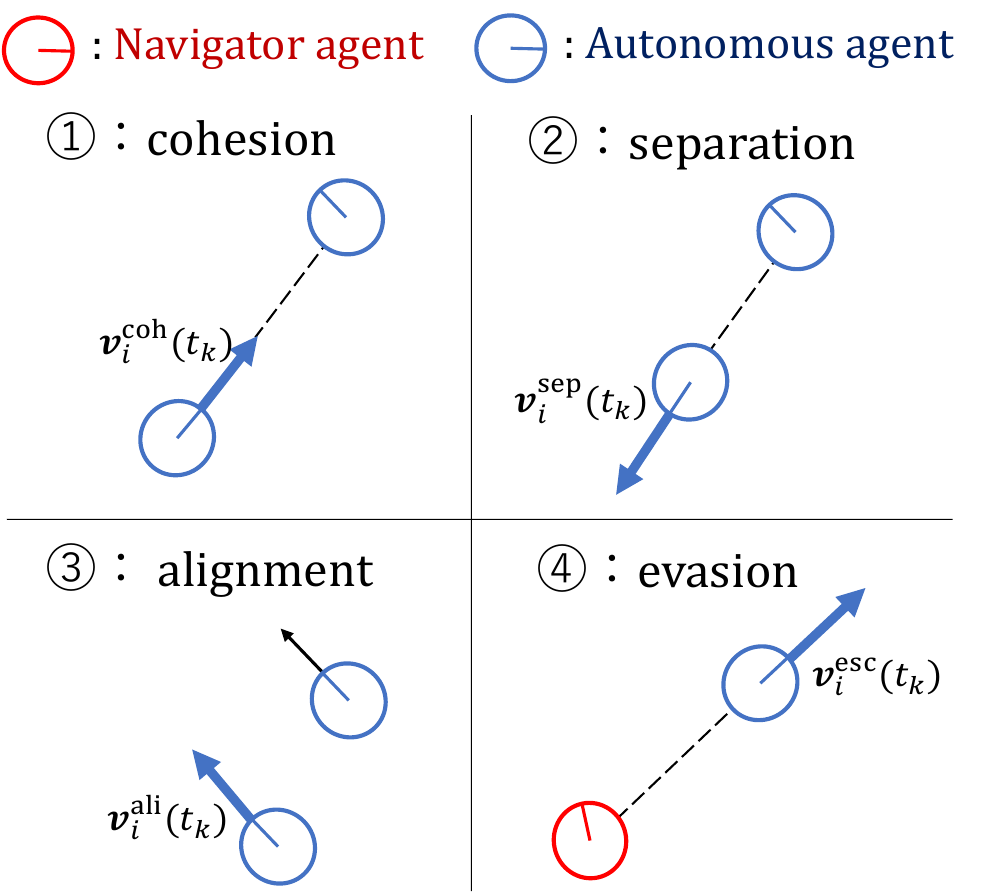}
    \caption{Conceptual diagram of the autonomous agent model inspired by the Boids model}
    \label{fig:boids}
\end{figure}

\begin{equation}
\label{eq:obs}
\begin{aligned}
&\mathcal{I} = \{1,\ldots,N\}\cup\{d\},\\[1mm]
&(\bm{p}_k(t_k),\bm{v}_k(t_k)) =
\begin{cases}
(\bm{p}_{s,k}(t_k),\bm{v}_{s,k}(t_k)), \\\qquad (k=1,\ldots,N),\\
(\bm{p}_{d}(t_k),\bm{v}_{d}(t_k)), \\\qquad (k=d),
\end{cases}\\[1mm]
&R_k =
\begin{cases}
R_s, & (k=1,\ldots,N),\\
R_d, & (k=d),
\end{cases}\\[1mm]
&\mathcal{N}_k(t_k)
= \left\{\, \ell\in\mathcal{I}\setminus\{k\}\ \middle|\ \|\bm{p}_\ell(t_k)-\bm{p}_k(t_k)\|\le R_k \right\},\\[1mm]
&\mathcal{O}_k(t_k)
= \left\{\, (\bm{p}_\ell(t_k),\bm{v}_\ell(t_k))\ \middle|\ \ell\in\mathcal{N}_k(t_k) \right\},\\[1mm]
&\bm{a}_{s,i}(t_k) = \pi_s\!\left(\bm{p}_{s,i}(t_k),\bm{v}_{s,i}(t_k),\mathcal{O}_i(t_k)\right),\\&\qquad i=1,\ldots,N,\\
&\bm{a}_{d}(t_k) = \pi_d\!\left(\bm{p}_d(t_k),\bm{v}_d(t_k),\mathcal{O}_d(t_k),\bm{p}_{\mathrm{goal}}\right).
\end{aligned}
\end{equation}
where, in eq.~(\ref{eq:obs}), $\mathcal{I}$ denotes the index set of all agents, $\mathcal{N}_k(t_k)$ denotes the index set of agents observed by agent $k$ at time $t_k$,
and $\mathcal{O}_k(t_k)$ denotes the set of information observed by agent $k$ at time $t_k$.

\section{Proposal of autonomous agent model}
\label{sec:sheep}

As a representative motion model for autonomous agents that has been widely used in the literature, a model based on the Boids model~\cite{boids} is well known as shown in Fig.~\ref{fig:boids}.
It is composed of the following four interaction rules:
\begin{enumerate}
	\item[1.] \textbf{Cohesion:} nearby agents are attracted to each other.
	\item[2.] \textbf{Separation:} nearby agents keep distance to avoid collisions.
	\item[3.] \textbf{Alignment:} nearby agents match their velocities.
	\item[4.] \textbf{Evasion:} agents evade the navigator agent.
\end{enumerate}
On the other hand, there also exist models that omit the alignment term, in which autonomous agents do not reference the velocity information of surrounding agents and instead move based solely on relative positional relationships~\cite{Yusuke_Tsunoda2021}.
In the former class of models, the alignment term allows motion to propagate rapidly through the swarm, making it easier for the entire group to translate as a coherent body.
However, depending on the specific formulation of alignment, a residual propulsive component may remain even in situations where the swarm should come to rest, which can cause undesired drifting and necessitate re-collection of the swarm.
In contrast, since the latter class of models does not use velocity information, drifting tends to be suppressed compared to models with alignment; nevertheless, because the interaction structure depends only on relative positions, the ``intention to move'' can be difficult to propagate across the swarm, potentially resulting in behaviors where the swarm becomes excessively tightly bound and less responsive as a group.

Based on these considerations, the desirable swarm behavior as a guidance target can be summarized as follows:
\begin{enumerate}
    \item[1.] When the navigator agent intervenes, the induced motion propagates within the swarm so that the swarm moves as a cohesive unit.
    \item[2.] When the intervention becomes weak, no excessive propulsive component remains, and the swarm tends toward rest.
\end{enumerate}
Therefore, this study constructs a \emph{reference-velocity} model for autonomous agents that simultaneously achieves
(i) coherent collective motion through motion propagation within the swarm and
(ii) convergence toward rest without residual propulsion when the intervention of the navigator agent weakens.

Specifically, we define the reference velocity of autonomous agent $i$,
$\bm{v}_{s,i}^{\mathrm{ref}}(t_k)\in\mathbb{R}^{3}\,[\mathrm{m/s}]$,
as the weighted sum of four interaction terms corresponding to cohesion, separation, alignment, and evasion, as follows:

\begin{equation}
    \begin{aligned}
        \bm{v}_{s,i}^{\mathrm{ref}}(t_k)
        = &w_{\mathrm{coh}}\bm{v}^{\mathrm{coh}}_{i}(t_k)
        + w_{\mathrm{sep}}\bm{v}^{\mathrm{sep}}_{i}(t_k)\\
        &+ w_{\mathrm{ali}}\bm{v}^{\mathrm{ali}}_{i}(t_k)
        + w_{\mathrm{esc}}\bm{v}^{\mathrm{esc}}_{i}(t_k)
        \label{eq:vsref_def},
    \end{aligned}
\end{equation}
where $t_k$ denotes the discrete time, and $w_{\mathrm{coh}}, w_{\mathrm{sep}}, w_{\mathrm{ali}}, w_{\mathrm{esc}} \in \mathbb{R}_+$ are the weights for the respective terms.
Each interaction term is computed solely from the information $\mathcal{O}_i(t_k)$ of agents contained in the observation set $\mathcal{N}_{i}(t_k)$ defined in Section~\ref{sec:problem}.
In what follows, we decompose the observation set $\mathcal{N}_{i}(t_k)$ into the sets of observed autonomous agents and observed navigator agent(s) as
\begin{equation}\label{eq:neighbor_split}
    \begin{split}
\mathcal{N}^{s}_{i}(t_k) = \{\, j\in \mathcal{N}_{i}(t_k)\mid j \text{ is an \emph{autonomous agent}}\},\\
\mathcal{N}^{d}_{i}(t_k) = \{\, j\in \mathcal{N}_{i}(t_k)\mid j \text{ is a \emph{navigator agent}}\}.
    \end{split}
\end{equation}

\paragraph{Distance Correction for Collision Avoidance}
The separation term plays a crucial role in collision avoidance.
In an idealized setting where agents can track arbitrary velocities and are not subject to acceleration limits, non-collision can be guaranteed by defining the separation term as a repulsive force inversely proportional to the inter-agent distance.
However, in the present study, since each agent is subject to velocity and acceleration constraints, providing a separation term that prevents collisions in the ideal setting does not necessarily ensure that a sufficient evasive maneuver can be realized.

To enhance safety under such constraints, we inflate each agent using a safety radius $r_{\mathrm{safe}}\in\mathbb{R}_+\,[\mathrm{m}]$ that satisfies $r_{\mathrm{indiv}}<r_{\mathrm{safe}}$, and we redefine the inter-agent distance as the minimum distance after inflation.
Let the center-to-center distance between autonomous agents $i$ and $j$ be
\begin{equation}\label{eq:dist}
d_{ij}(t_k) = \|\bm{p}_{s,i}(t_k)-\bm{p}_{s,j}(t_k)\|,
\end{equation}
and define the inflated distance that accounts for the agent radii as
\begin{equation}\label{eq:infl_dist}
h_{ij}(t_k) = d_{ij}(t_k)-2r_{\mathrm{safe}}.
\end{equation}
When the separation term is specified in an inversely proportional form, using $h_{ij}(t_k)$ directly may cause the separation term to diverge as the distance approaches zero near imminent collision.
Therefore, we smooth the distance as
\begin{equation} \label{eq:infl_smooth}
\bar d_{ij}(t_k) = \sqrt{\max\!\bigl(h_{ij}(t_k),0\bigr)^{2}+\delta^{2}}.
\end{equation}
Here, $\delta>0$ is a smoothing parameter.
With this definition, even when the safety regions of agents overlap, $\bar d_{ij}(t_k)$ is lower-bounded by $\delta$ and thus never becomes zero.

\paragraph{Cohesion Term}
The cohesion term encourages autonomous agents to aggregate.
Let $\mathcal{N}^s_i(t_k)$ denote the set of neighboring autonomous agents of autonomous agent $i$ at time $t_k$, and let $|\mathcal{N}^s_i(t_k)|$ be its cardinality.
We define the mean position of the neighbors as
\begin{equation}
\bar{\bm{p}}_{s,i}(t_k)=
\begin{cases}
\dfrac{1}{| \mathcal{N}^s_i(t_k) |}\displaystyle\sum_{j\in \mathcal{N}^s_i(t_k)} \bm{p}_{s,j}(t_k), & |\mathcal{N}^s_i(t_k)|>0,\\[6pt]
\bm{p}_{s,i}(t_k), & |\mathcal{N}^s_i(t_k)|=0,
\end{cases}
\end{equation}
and then specify the cohesion term as
\begin{equation}
\bm v^{\mathrm{coh}}_{i}(t_k)=\bar{\bm{p}}_{s,i}(t_k)-\bm{p}_{s,i}(t_k).
\label{eq:v_coh}
\end{equation}

\paragraph{Separation Term}
The separation term encourages nearby autonomous agents to maintain a safe distance, thereby contributing to collision avoidance and inter-agent spacing.
Using the center-to-center distance $d_{ij}(t_k)$ between autonomous agents $i$ and $j$ defined in~(\ref{eq:dist}), we define the unit vector from agent $j$ to agent $i$ as
\begin{equation}
\bm{e}_{ij}(t_k)=
\frac{\bm p_{s,i}(t_k)-\bm p_{s,j}(t_k)}{d_{ij}(t_k)}.
\end{equation}
Using the smoothed inflated distance $\bar d_{ij}(t_k)$ defined in~(\ref{eq:infl_smooth}), we then define the separation term as
\begin{equation}
\bm{v}^{\mathrm{sep}}_{i}(t_k)=\sum_{j\in \mathcal{N}_i^{s}(t_k)} \frac{\bm{e}_{ij}(t_k)}{\bar d_{ij}(t_k)^{2}} .
\label{eq:v_sep}
\end{equation}
Since $\bar d_{ij}(t_k)$ is lower-bounded by $\delta$ even when the safety regions overlap, the separation term does not diverge.
Moreover, the inverse-square dependence $1/\bar d_{ij}(t_k)^{2}$ yields a local interaction: it induces strong avoidance at short range, while its influence decays rapidly at longer distances.

\paragraph{Alignment Term}
The alignment term plays the role of propagating the motion induced by the navigator agent throughout the swarm.
By defining it as a function of the mean velocity that includes a virtual neighboring agent with zero velocity, we endow the model with the property that the swarm velocity naturally decays when the intervention from the navigator agent becomes weak.

As in the cohesion term, let $\mathcal{N}^{s}_{i}(t_k)$ be the set of neighboring autonomous agents observed by autonomous agent $i$, and let $|\mathcal{N}^{s}_{i}(t_k)|$ be its cardinality.
Then, the alignment term is given as follows:

\begin{equation}
\bm v^{\mathrm{ali}}_{i}(t_k)
=
\frac{
\bm v_{s,i}(t_k) + \sum\limits_{j \in \mathcal{N}^{s}_{i}(t_k)} \bm v_{s,j}(t_k)
}{
|\mathcal{N}^{s}_{i}(t_k)| + 1 + w_{\mathrm{v}}
}.
\end{equation}
Here, $w_{\mathrm{v}}\in\mathbb{R}_+$ represents the ``number of virtual neighboring agents with zero velocity.''
By including the agent's own velocity $\bm v_{s,i}(t_k)$ in the average and setting $w_{\mathrm{v}}>0$, we obtain
$\bm v^{\mathrm{ali}}_{i}(t_k)=\bm v_{s,i}(t_k)/(1+w_{\mathrm{v}})$ even when no neighboring agents are present.
Hence, the alignment term acts in a direction that attenuates the velocity.
As a result, this formulation suppresses the phenomenon that the swarm continues to drift due to inertia when the intervention from the navigator agent is weak.

\begin{figure}[t]
    \centering
    \includegraphics[width=\linewidth]{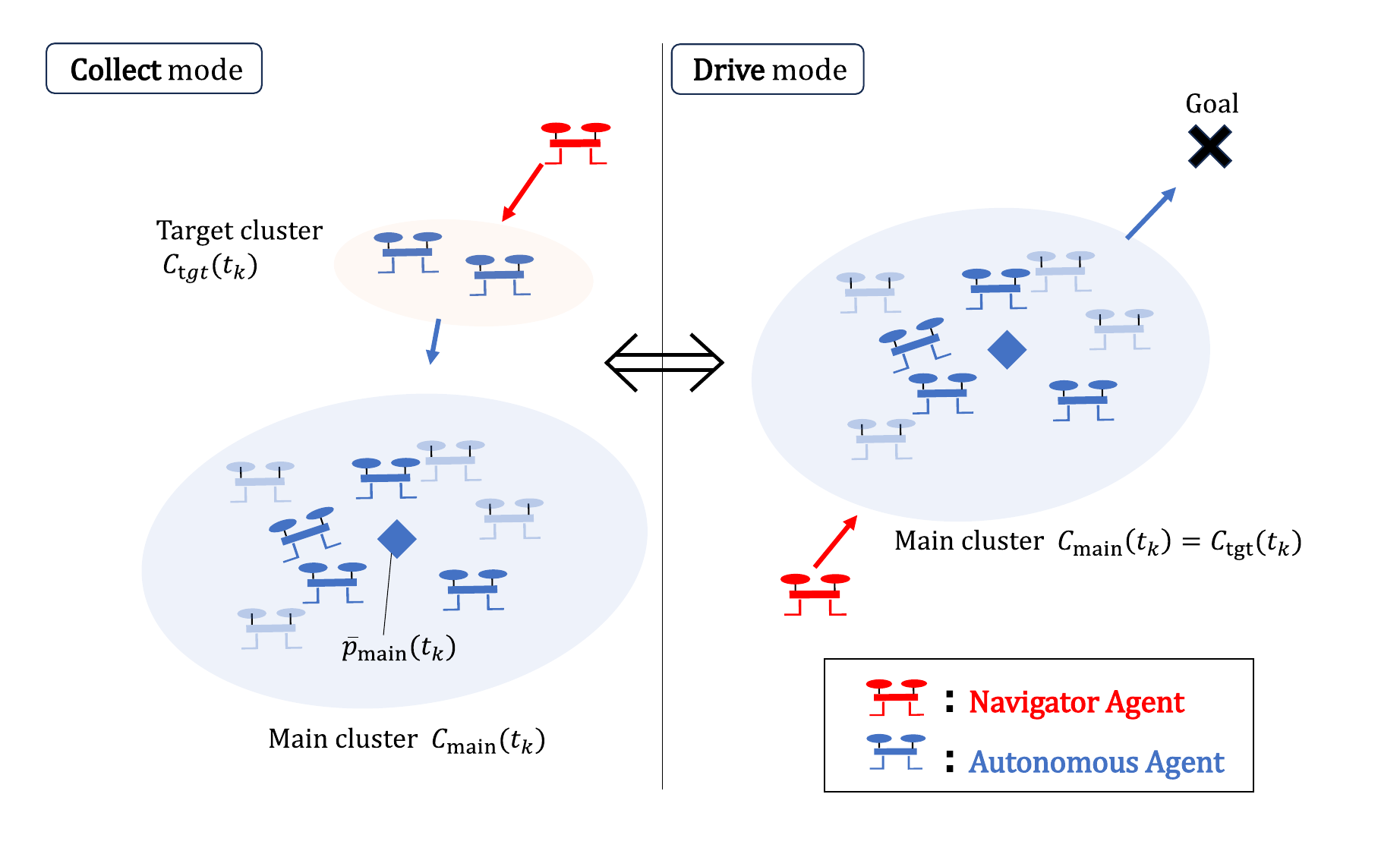}
    \caption{Conceptual diagram of collecting and driving}
    \label{fig:collect_drive}
\end{figure}

\paragraph{Evasion Term}
The evasion term functions as an external force exerted by the navigator agent on the autonomous agents and constitutes the primary driving mechanism induced by the intervention of the navigator agent.
Let $\mathcal{N}^{d}_{i}(t_k)$ denote the set of navigator agent(s) observed by autonomous agent $i$.
Define the relative position between agent $i$ and a navigator agent $d$ as
\begin{equation}
\bar d_{id}(t_k)=\sqrt{\max\!\left(d_{id}(t_k)-2r_{\mathrm{safe}},\,0\right)^{2}+\delta^{2}}.
\end{equation}
Then, the evasion term is given by
\begin{equation}
\bm v^{\mathrm{esc}}_{i}(t_k)=\sum_{d \in \mathcal{N}^{d}_{i}(t_k)}
\frac{\bm r_{id}(t_k)}{d_{id}(t_k)}\cdot\frac{1}{\bar d_{id}(t_k)^{2}}.
\end{equation}
This term is a repulsive interaction obtained by weighting the unit direction vector pointing away from the navigator agent by the inverse square of the modified distance.
The inflation together with the smoothing parameter $\delta$ prevents excessively large inputs at extremely short distances while still providing sufficiently strong avoidance responses when agents approach each other.
When the navigator agent moves outside the sensing range, $\mathcal{N}^{d}_{i}(t_k)$ becomes the empty set and the velocity component induced by the evasion term becomes zero.
Any residual velocity then naturally decays through the alignment term, which helps ensure that the swarm comes to rest near the goal and/or under non-intervention conditions.

\begin{figure}[t]
    \centering
    \includegraphics[width=0.9\linewidth]{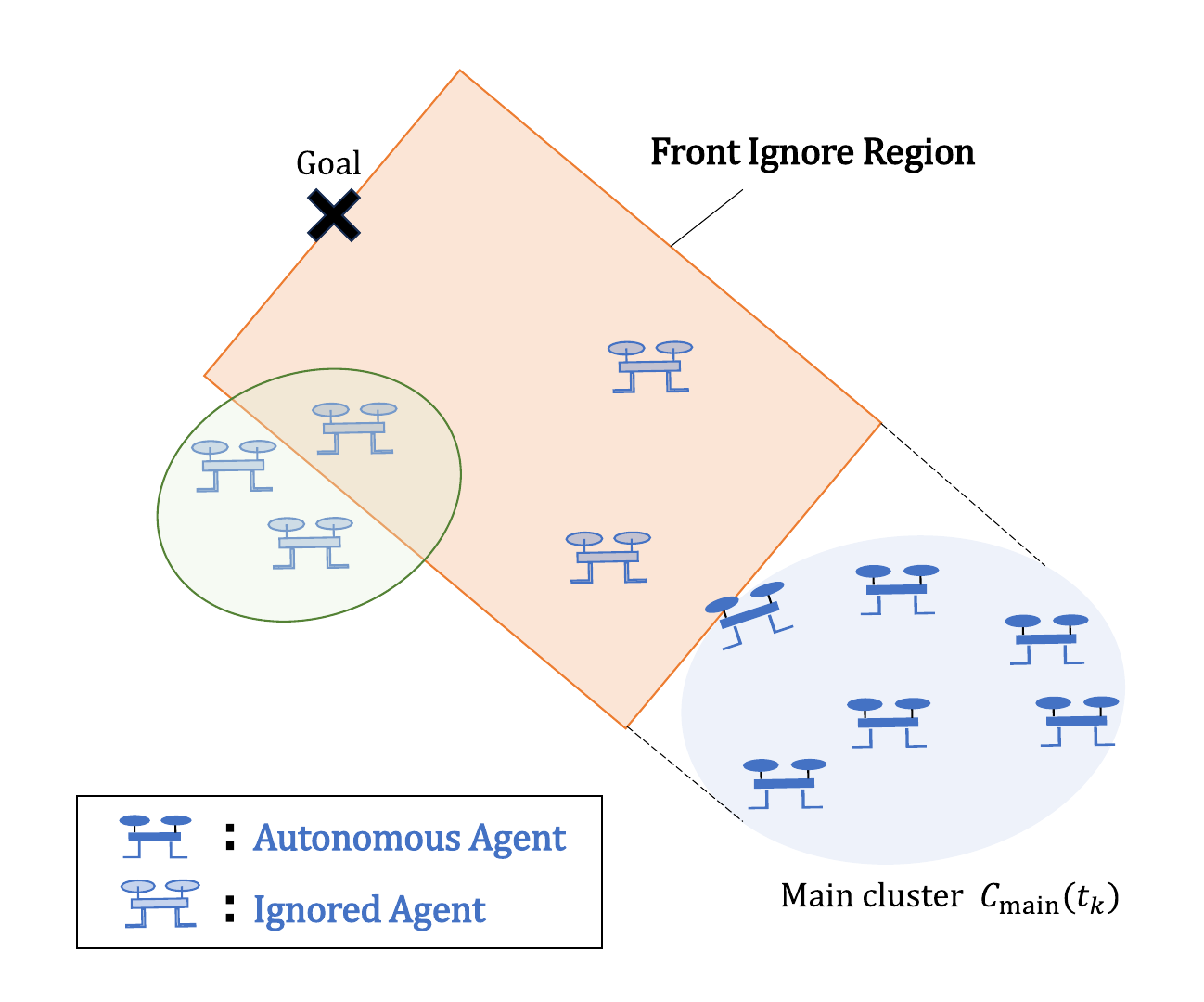}
    \caption{Conceptual diagram of ignore region}
    \label{fig:ignore}
\end{figure}

\section{Proposal of navigator agent controller}
\label{sec:shepherd}
In this section, we design the input acceleration $\bm{a}_d(t_k)$ of the navigator agent.
Since the autonomous agents do not necessarily form a single connected group and may split into multiple aggregates (clusters), it is necessary to define a strategy that determines (i) which group should be treated as the guidance target and (ii) toward which direction that group should be driven.
To this end, we introduce clustering based on DBSCAN~\cite{DBSCAN} and switch between a \emph{collection} strategy and a \emph{guidance} strategy.
Moreover, if the navigator agent relies only on instantaneous geometric relationships, the resulting guidance law may fail to track continuously varying velocities and swarm shapes, potentially leading to inefficient behavior.
Inspired by the Dynamic Window Approach (DWA)~\cite{dwa}, we adopt a framework that generates motion candidates under constraints, predicts the short-horizon outcome, and selects an action based on an evaluation function.
Specifically, the navigator agent generates feasible motion candidates within its constraints, evaluates them according to multiple criteria, and selects the motion that is considered optimal.

\subsection{Selection of the guidance target by clustering}
Fig.~\ref{fig:collect_drive} shows an overview diagram of the guidance strategy for the guidance agent.
At every fixed period $T_{\mathrm{period}}$~[s], the navigator agent clusters the observed autonomous agents using DBSCAN.
When multiple clusters exist and the inter-cluster separation is sufficiently large, the navigator agent selects the largest cluster as the main cluster $C_{\mathrm{main}}(t_k)$ and selects the cluster farthest from the main cluster as the target cluster $C_{\mathrm{tgt}}(t_k)$.
It then switches to the \emph{collection mode}, in which the target cluster is collected (merged) into the main cluster.
In contrast, when only a single cluster exists, or when all clusters other than the main cluster exist only along the path toward the goal (Fig.~\ref{fig:ignore}), the navigator agent sets the target cluster to be the main cluster (i.e., $C_{\mathrm{main}}(t_k)=C_{\mathrm{tgt}}(t_k)$) and switches to the \emph{guidance mode}, in which the swarm is steered toward the goal position.

Furthermore, in the collection mode, to prevent mode chattering caused by changes in the cluster structure, we record the autonomous agents belonging to the main cluster and the target cluster at the moment when the collection mode starts.
Until the collection process terminates, at each clustering cycle, we define the main cluster and the target cluster as the clusters that contain the largest number of the recorded agents, respectively.
In addition, to account for the possibility that the navigator agent may temporarily lose observation of the main cluster during collection, we maintain a record of the main-cluster position so that guidance can be continued even when the main cluster is out of the sensing range.

\subsection{Motion selection of the navigator agent}
\begin{figure}[t]
    \centering
    \includegraphics[width=0.7\linewidth]{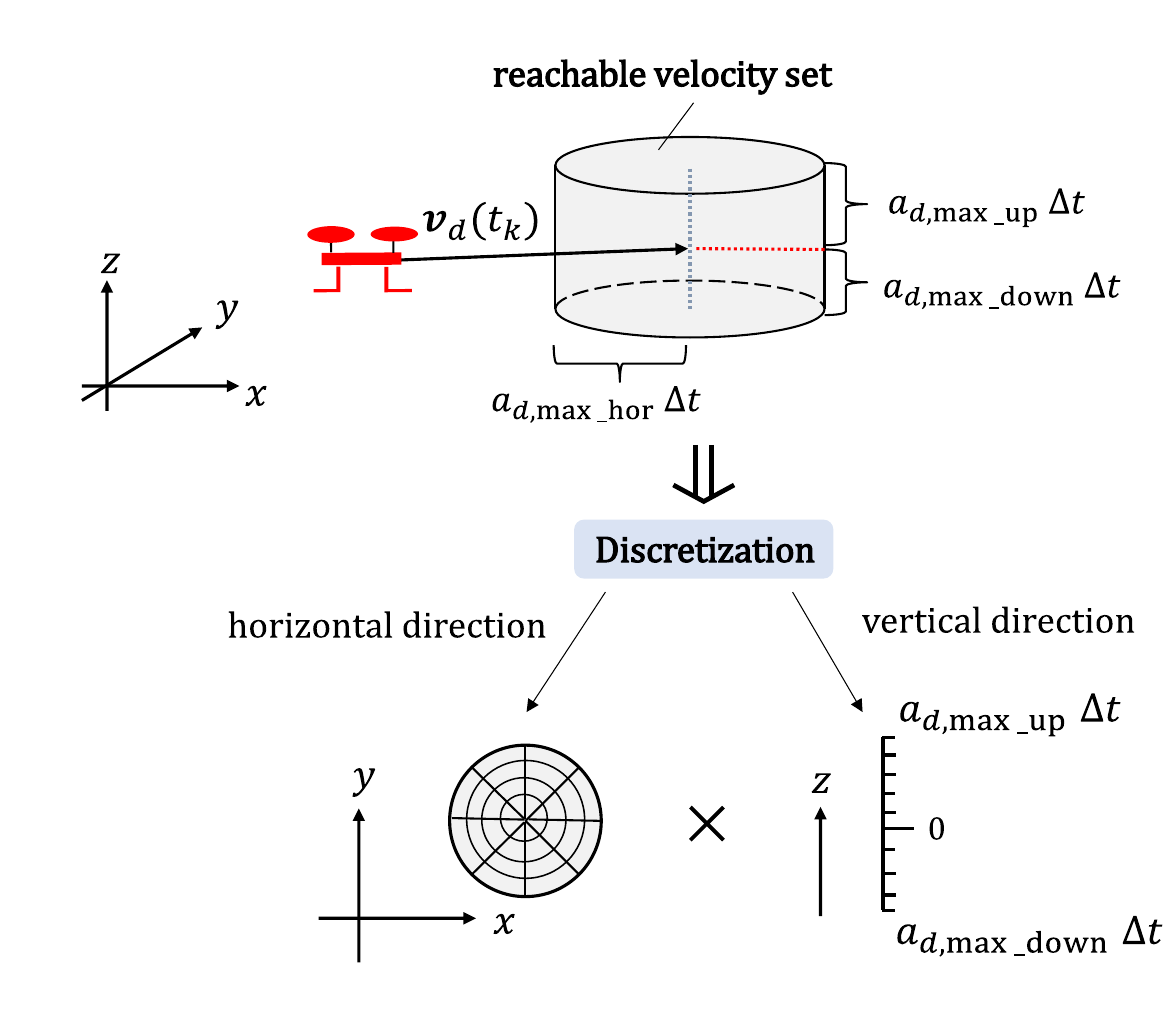}
    \caption{Conceptual diagram of acceleration candidate generation}
    \label{fig:dwa_candidate}
\end{figure}

In this study, we select the input $\bm{a}_d(t_k)$ from feasible motion candidates under constraints using a DWA-inspired framework of ``candidate generation--short-horizon prediction--evaluation and selection.''
For the admissible accelerations, we discretize the horizontal component in terms of its heading and magnitude, and discretize the vertical component in terms of its magnitude.
For each sampled acceleration $\bm{a}^{(m)}_d$ obtained from the Cartesian product of these discretizations, we also discretize the length of the accelerating segment along the predicted trajectory, denoted by $N_{\mathrm{acc}}\in\mathbb{R}_+$.
In this way, we generate a finite set of motion candidates over the prediction horizon $T_{\mathrm{pred}}\in\mathbb{R}_+$~[s].
Figure~\ref{fig:dwa_candidate} illustrates the concept of candidate generation.

For each generated motion candidate, we predict the response of the autonomous swarm over the horizon $T_{\mathrm{pred}}$ using the autonomous-agent model.
As with clustering, this prediction is performed every $T_{\mathrm{period}}$.

Each candidate is evaluated from two perspectives: \emph{terminal evaluation} and \emph{path evaluation}.
The evaluation consists of the following criteria.

\paragraph{Terminal evaluation}
\begin{itemize}
    \item \textbf{Velocity cost} $c^{(m)}_{\mathrm{vel}}$: the speed and heading of the target cluster.
    \item \textbf{Positioning cost} $c^{(m)}_{\mathrm{pos}}$: the relative positioning of the navigator agent with respect to the target cluster.
    \item \textbf{Observation-maintenance cost} $c^{(m)}_{\mathrm{obs}}$: the risk of losing observation.
\end{itemize}

\paragraph{Path evaluation}
\begin{itemize}
    \item \textbf{Split-avoidance cost} $c^{(m)}_{\mathrm{split}}$: the risk of inducing fragmentation (splitting).
\end{itemize}
We compute the weighted sum of these criteria (see~(\ref{cost})) and adopt the candidate that yields the best (i.e., smallest) evaluation value:
\begin{equation}
    \begin{aligned}
        J^{(m)}
        =& w_{\mathrm{vel}} c^{(m)}_{\mathrm{vel}}
        + w_{\mathrm{pos}} c^{(m)}_{\mathrm{pos}}
        + w_{\mathrm{obs}} c^{(m)}_{\mathrm{obs}} \\
        &+ w_{\mathrm{split}} c^{(m)}_{\mathrm{split}}
        + w_{\mathrm{clear}} c^{(m)}_{\mathrm{clear}}
        + w_{\mathrm{alt}} c^{(m)}_{\mathrm{alt}}.
        \label{cost}
    \end{aligned}
\end{equation}
Here, $w_{\mathrm{vel}}, w_{\mathrm{pos}}, w_{\mathrm{obs}}, w_{\mathrm{split}}, w_{\mathrm{clear}}, w_{\mathrm{alt}} \in \mathbb{R}_+$ are the weights for the respective evaluation criteria.

\subsubsection{Velocity evaluation}
\begin{figure}[t]
    \centering
    \includegraphics[width=0.9\linewidth]{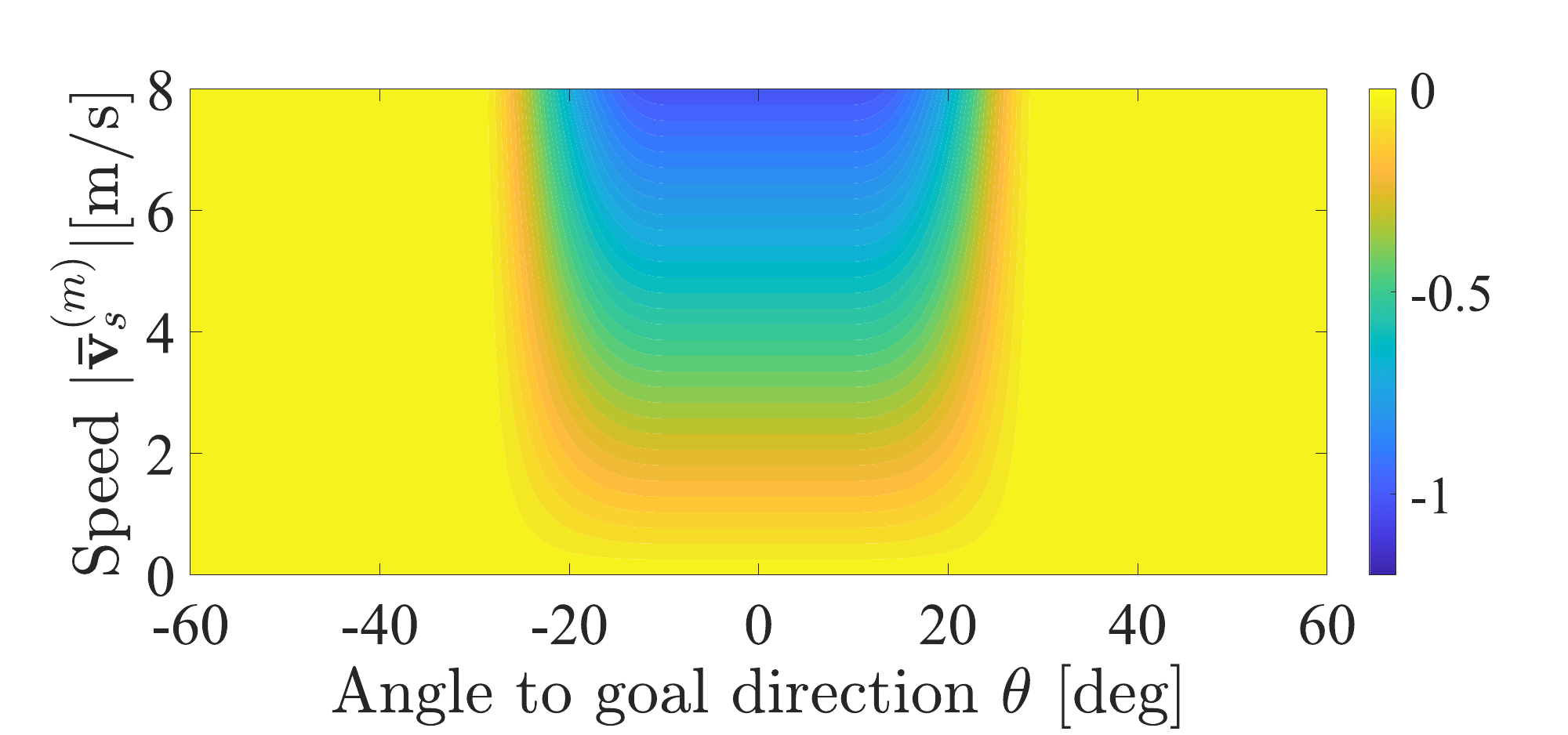}
    \caption{Heat map of velocity evaluation}
    \label{fig:vel_eval}
\end{figure}

In the velocity evaluation, we compute the mean velocity of the target cluster at the end of the prediction horizon, denoted by $\bar{\bm{v}}_{\mathrm{s}}^{(m)}$, and assign a better score when the velocity component toward the goal direction is larger.
The evaluation is designed to decay as the cluster approaches the goal and to be ignored when the heading deviation exceeds a prescribed threshold.
The velocity-evaluation cost is defined as
\begin{equation}
    \label{eq:cost_vel}
    c_{\mathrm{vel}}^{(m)} = -\,\gamma^{(m)}\,v_{\mathrm{scale}}^{(m)}\,\eta^{(m)}.
\end{equation}
Here, $\gamma^{(m)}$ is a coefficient determined by the velocity heading, $v_{\mathrm{scale}}^{(m)}$ is the normalized velocity component toward the goal direction, and $\eta^{(m)}$ is a scaling factor based on the distance to the goal.
Figure~\ref{fig:vel_eval} shows a heat map of the velocity evaluation.

\subsubsection{Position evaluation}
\begin{figure}[t]
    \centering
    \includegraphics[width=0.5\linewidth]{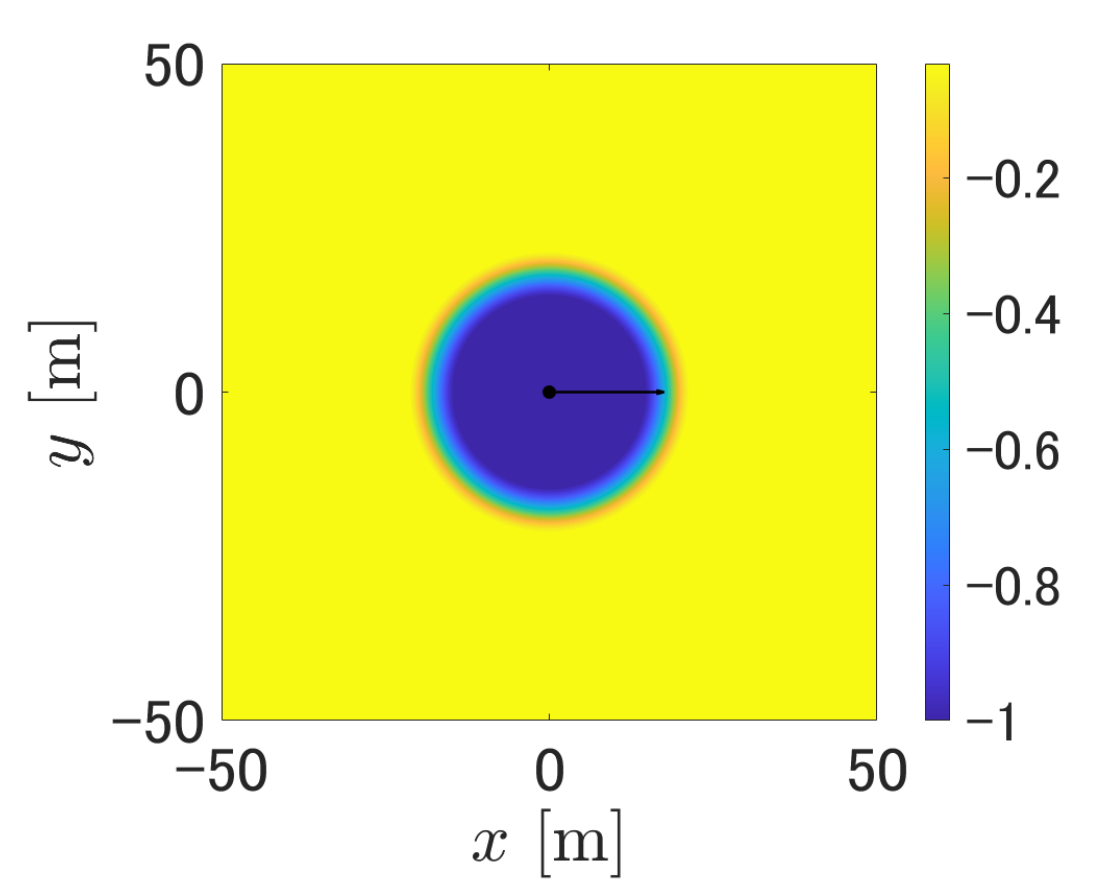}
    \includegraphics[width=0.5\linewidth]{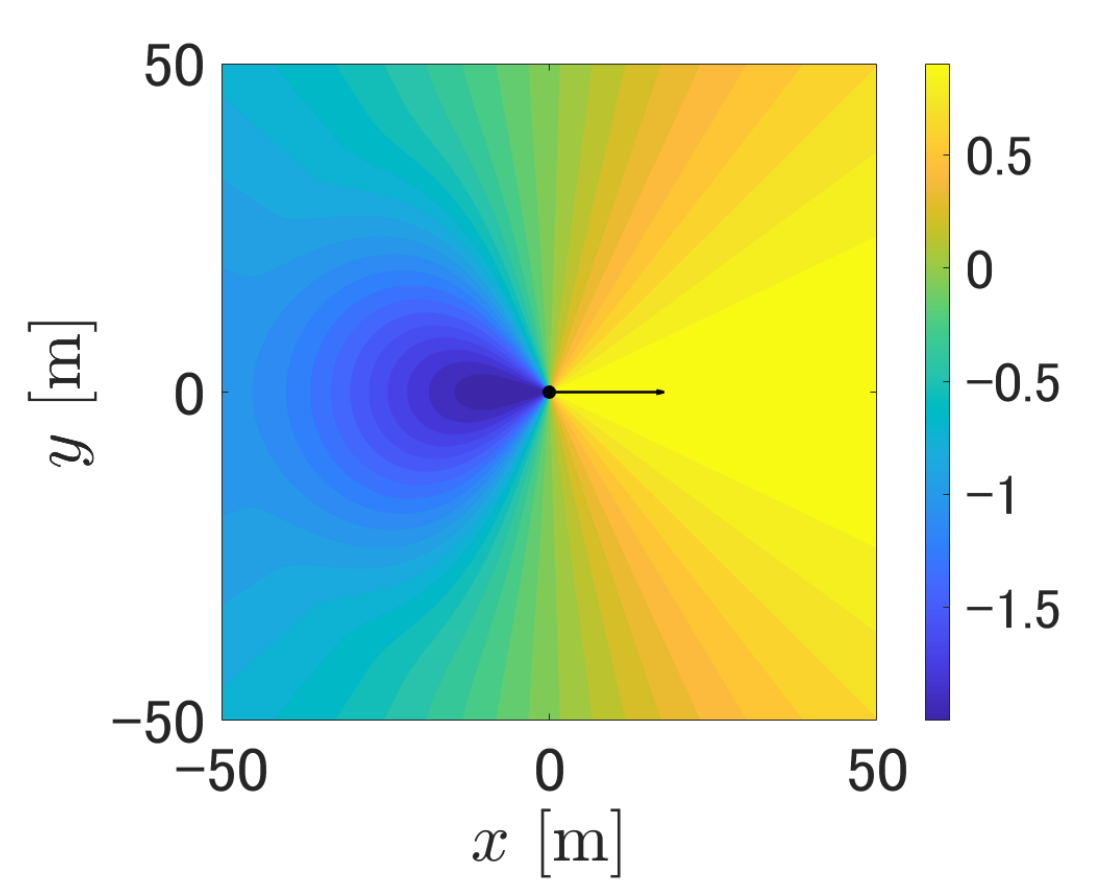}
    \caption{Heat map of position evaluation (top: far, bottom: near)}
    \label{fig:pos_eval}
\end{figure}

The position evaluation assesses the positioning of the navigator agent with respect to the target cluster.
The evaluation mainly depends on the angle and distance: when the navigator agent is initially close to the cluster, it favors a position behind the autonomous agents and at a short distance; when it is initially far, it emphasizes how much the navigator agent can approach the cluster over the horizon.
We smoothly switch between these regimes according to the initial distance.
The position-evaluation cost is defined as

\begin{equation}
    \label{eq:drivePot_def}
    c_{\mathrm{pos}}^{(m)}= -\Bigl((1-g_{\mathrm{far}})R_{\mathrm{align}}^{(m)} + g_{\mathrm{far}}R_{\mathrm{prog}}^{(m)}\Bigr).
\end{equation}
Here, $g_{\mathrm{far}}\in[0,1]$ is a weight determined by the initial distance, $R_{\mathrm{align}}^{(m)}$ is the near-range positioning score, and $R_{\mathrm{prog}}^{(m)}$ is the far-range positioning score.
Figure~\ref{fig:pos_eval} shows heat maps of the position evaluation for cases with far/near initial distances.
In the figure, the black dots represent autonomous agents belonging to the target cluster, and the arrows originating from them indicate the goal direction.

\subsubsection{Observation-maintenance evaluation}
The observation-maintenance evaluation penalizes a high risk of losing observation by focusing on the minimum distance between the navigator agent and the autonomous agents.
Let the threshold be defined as the observation radius $R_d$ minus a constant $\beta_{\mathrm{obs}}\in\mathbb{R}_+$, and let $d^{(m)}_{id\_\mathrm{min}}$ be the minimum distance between the navigator agent and the autonomous agents at the end of the horizon.
We define the cost as

\begin{equation}
    \begin{aligned}
        &\delta_{\mathrm{obs}}^{(m)}
        =
        \max\Bigl(0,\ d_{id\_{\min}}^{(m)}-(R_{d}-\beta_{\mathrm{obs}})\Bigr),\\
        &c_{\mathrm{obs}}^{(m)}
        = \delta_{\mathrm{obs}}^{(m)}\Bigl(1+ \delta_{\mathrm{obs}}^{(m)}\Bigr).
    \end{aligned}
\end{equation}
By including a quadratic term, this formulation penalizes more strongly when the distance exceeds the observation-maintenance threshold by a large margin, while the linear term ensures a non-negligible penalty even for minor violations.

\subsubsection{Split-avoidance evaluation}
\begin{figure}[t]
    \centering
    \includegraphics[width=0.8\linewidth]{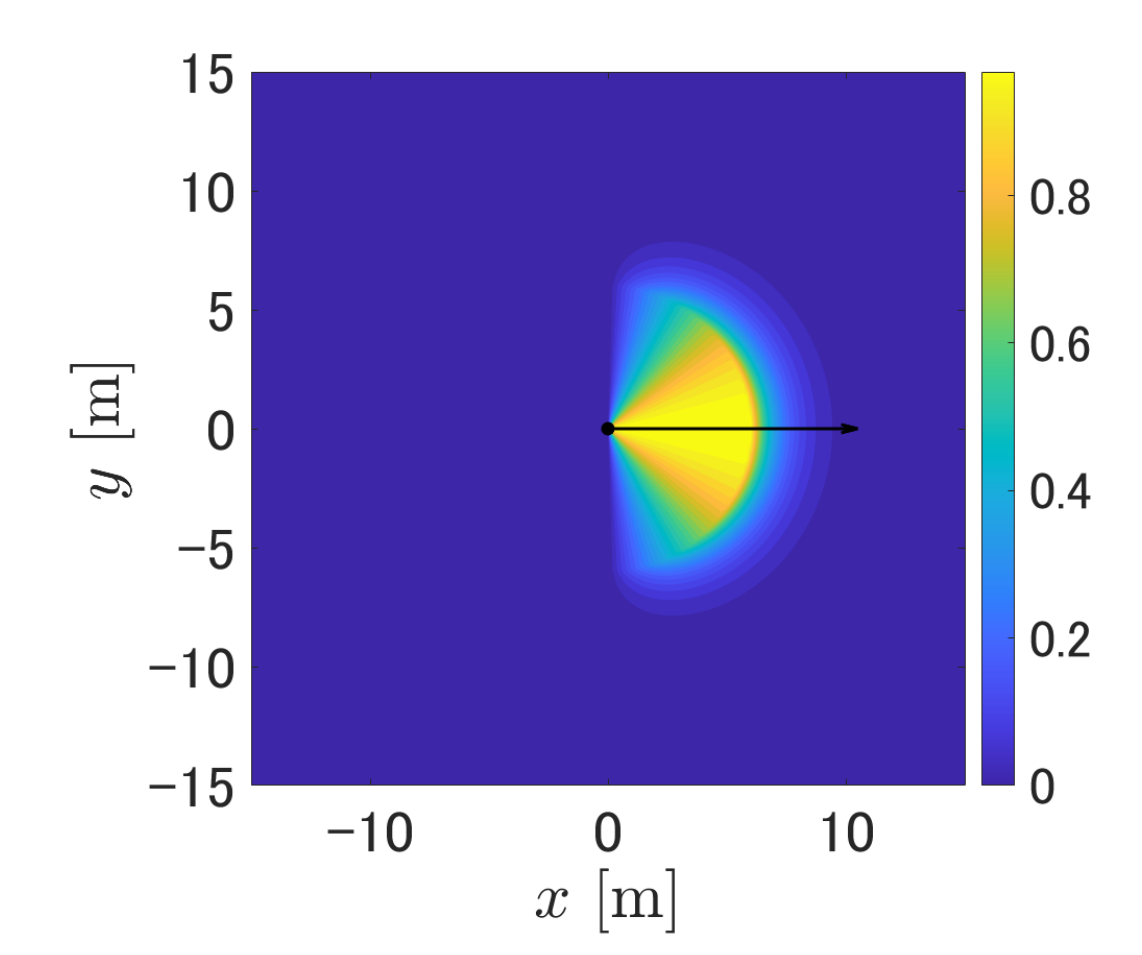}
    \caption{Heat map of split avoidance evaluation}
    \label{fig:split_eval}
\end{figure}

When moving to the rear of the swarm, if the navigator agent traverses across the interior of the swarm, the swarm may be fragmented.
Therefore, we penalize motion candidates that cross in front of the autonomous agents in the target cluster at close range.
We define the split-avoidance cost per step as
\begin{equation}
    \begin{aligned}
    \label{eq:avoid_step_cost}
    &c_{\mathrm{split}}^{(m)}(t_{k+j})\\
    &=
    \frac{1}{|C_{\mathrm{tgt}}|}
    \sum_{i\in C_{\mathrm{tgt}}}
    w_i^{(m)}(t_{k+j})\,
    b_{i\:\mathrm{split}}^{(m)}(t_{k+j})\,\xi_{i\:\mathrm{split}}^{(m)}(t_{k+j}),
    \end{aligned}
\end{equation}
where $C_{\mathrm{tgt}}$ is the set of agents in the target cluster and $|C_{\mathrm{tgt}}|$ is its cardinality.
Here, $w_i^{(m)}(t_{k+j})$ is a weight that increases as the distance between the navigator agent and autonomous agent $i$ decreases, $b_{i\:\mathrm{split}}^{(m)}(t_{k+j})$ is a weight that increases when the navigator agent is located in front of autonomous agent $i$, and $\xi_{i\:\mathrm{split}}^{(m)}(t_{k+j})$ is a coefficient that decreases as autonomous agent $i$ approaches the goal.
Figure~\ref{fig:split_eval} shows a heat map of the split-avoidance evaluation.

\subsubsection{Safety evaluation}

The safety evaluation considers both collision risk and altitude.
For collision-risk evaluation, we penalize cases where the minimum distance to any autonomous agent becomes smaller than $2r_{\mathrm{safe}}$.
For altitude evaluation, we penalize cases where the altitude of the navigator agent and/or autonomous agents falls below a prescribed threshold.
As in the observation-maintenance evaluation, we employ a combination of quadratic and linear terms.
Note that motion candidates that result in collision or violate the minimum-altitude constraint are excluded from evaluation.
The per-step costs for collision risk and altitude are defined by (\ref{eq:safe_step_cost}) and (\ref{eq:alt_step_cost}), respectively, and the final costs are taken as the maximum value over the prediction horizon:
\begin{equation}
    \label{eq:safe_step_cost}
    c_{\mathrm{clear}}^{(m)}(t_{k+j})
    =\delta_{\mathrm{clear}}^{(m)}(t_{k+j}) \Bigl(1 + \delta_{\mathrm{clear}}^{(m)}(t_{k+j}) \Bigr),
\end{equation}
\begin{equation}
    \label{eq:alt_step_cost}
    c_{\mathrm{alt}}^{(m)}(t_{k+j})
    =\delta_{\mathrm{alt}}^{(m)}(t_{k+j}) \Bigl(1 + \delta_{\mathrm{alt}}^{(m)}(t_{k+j}) \Bigr),
\end{equation}
where $\delta_{\mathrm{clear}}^{(m)}(t_{k+j})$ and $\delta_{\mathrm{alt}}^{(m)}(t_{k+j})$ denote the violation magnitudes associated with collision risk and altitude, respectively.

\section{Simulation Verification}
\label{sec:sim}
\begin{figure}[t]
    \centering
    \includegraphics[width=0.99\linewidth]{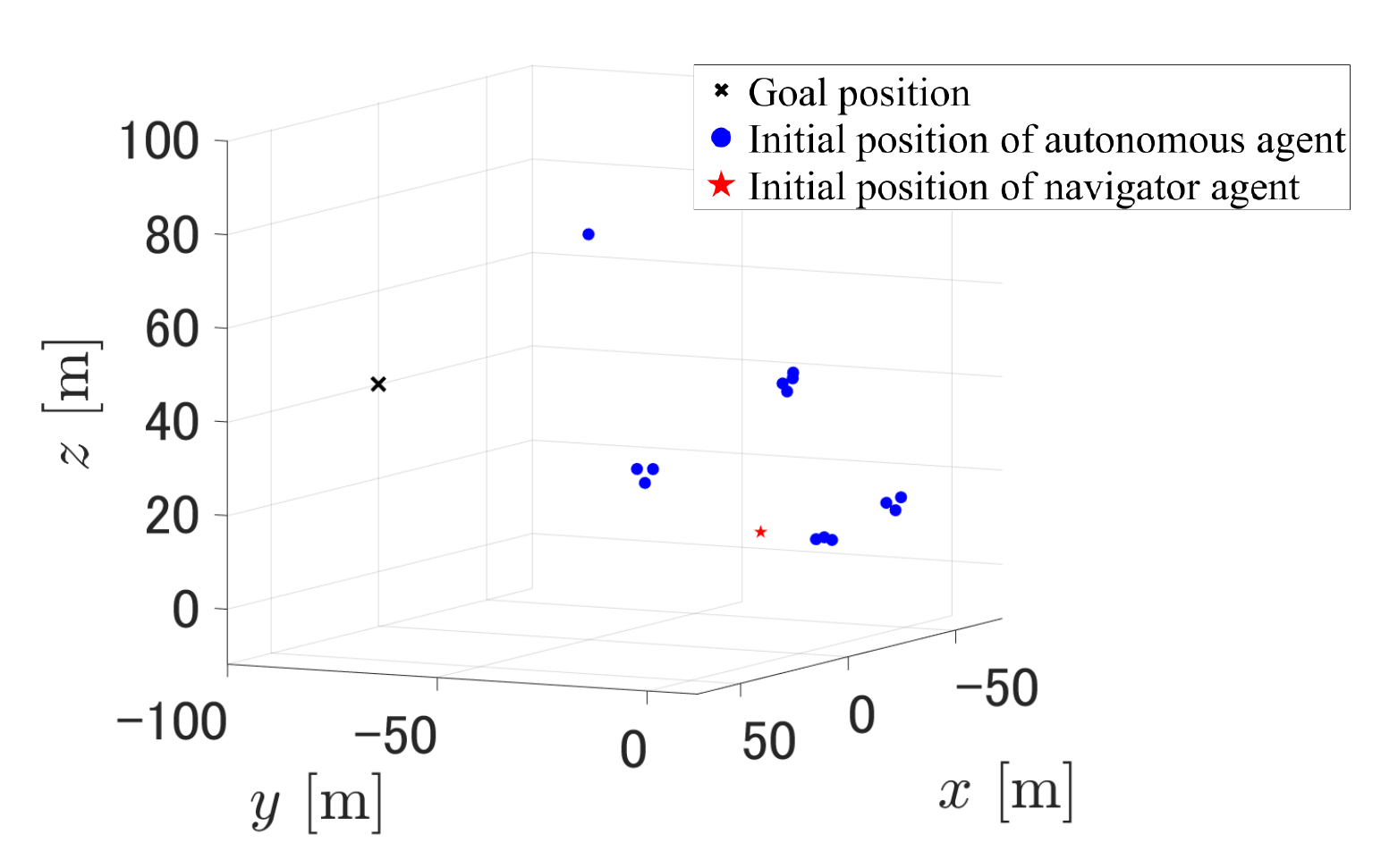}
    \caption{Initial positions of each agent}
    \label{fig:bunsan_initPos}
\end{figure}

\begin{figure}[t]
    \centering
    \includegraphics[width=0.9\linewidth]{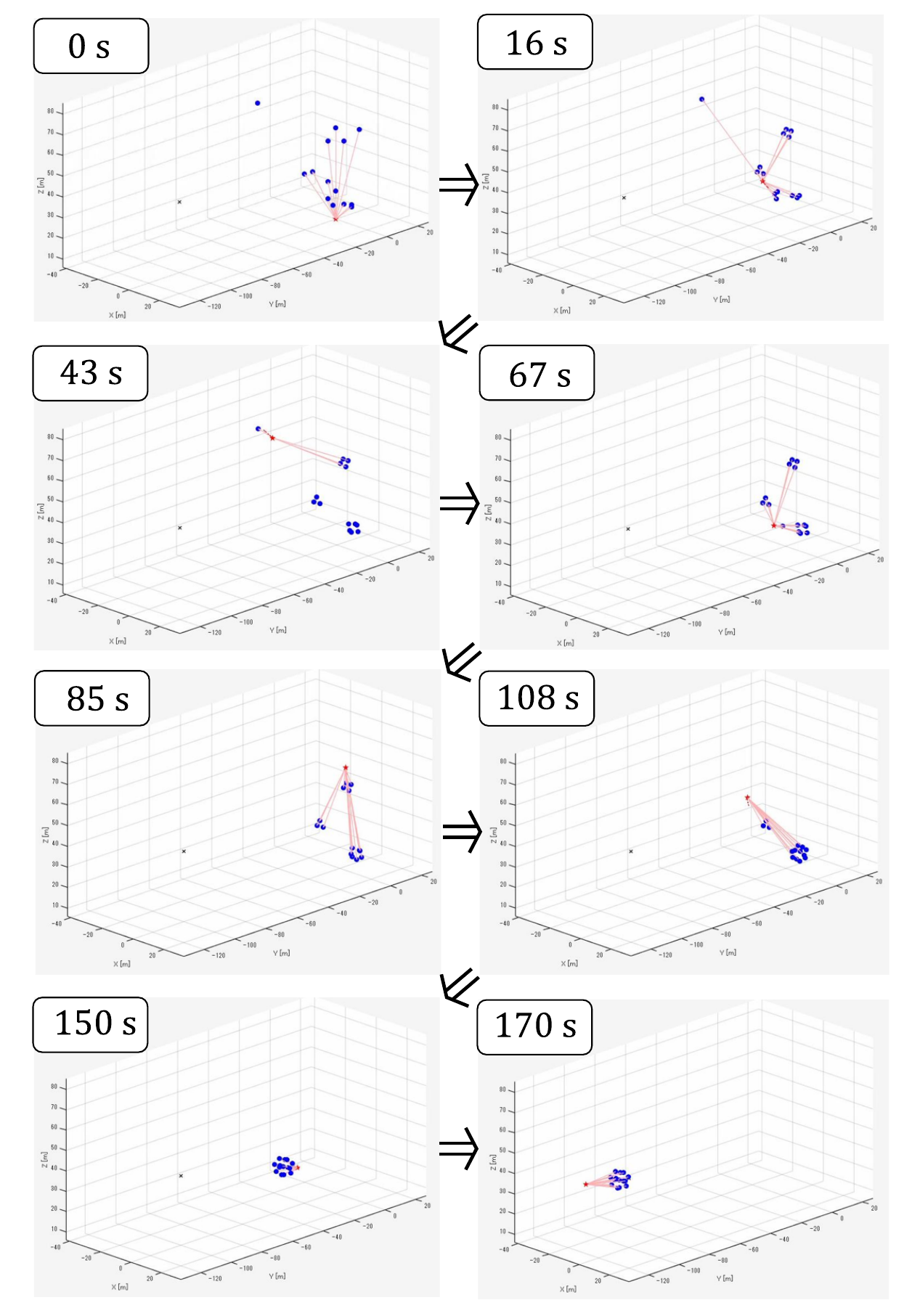}
    \caption{Snapshots of the guidance process}
    \label{fig:sim_result2_snap}
\end{figure}

\begin{figure}[t]
    \centering
    \includegraphics[width=0.9\linewidth]{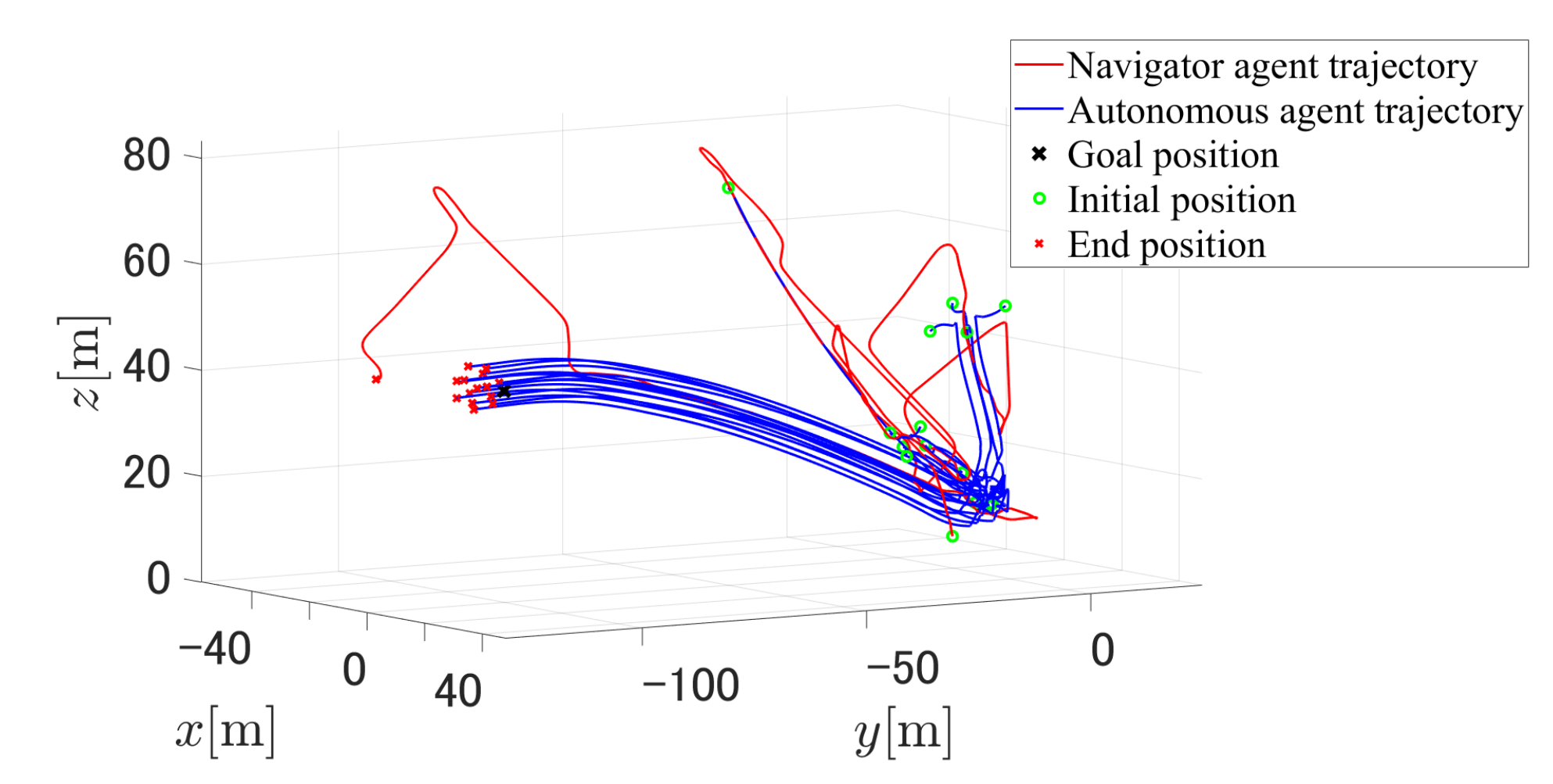}
    \caption{Simulation result (sparse initial positions)}
    \label{fig:sim_result_sparse}
\end{figure}

\begin{figure}[t]
    \centering
    \includegraphics[width=0.8\linewidth]{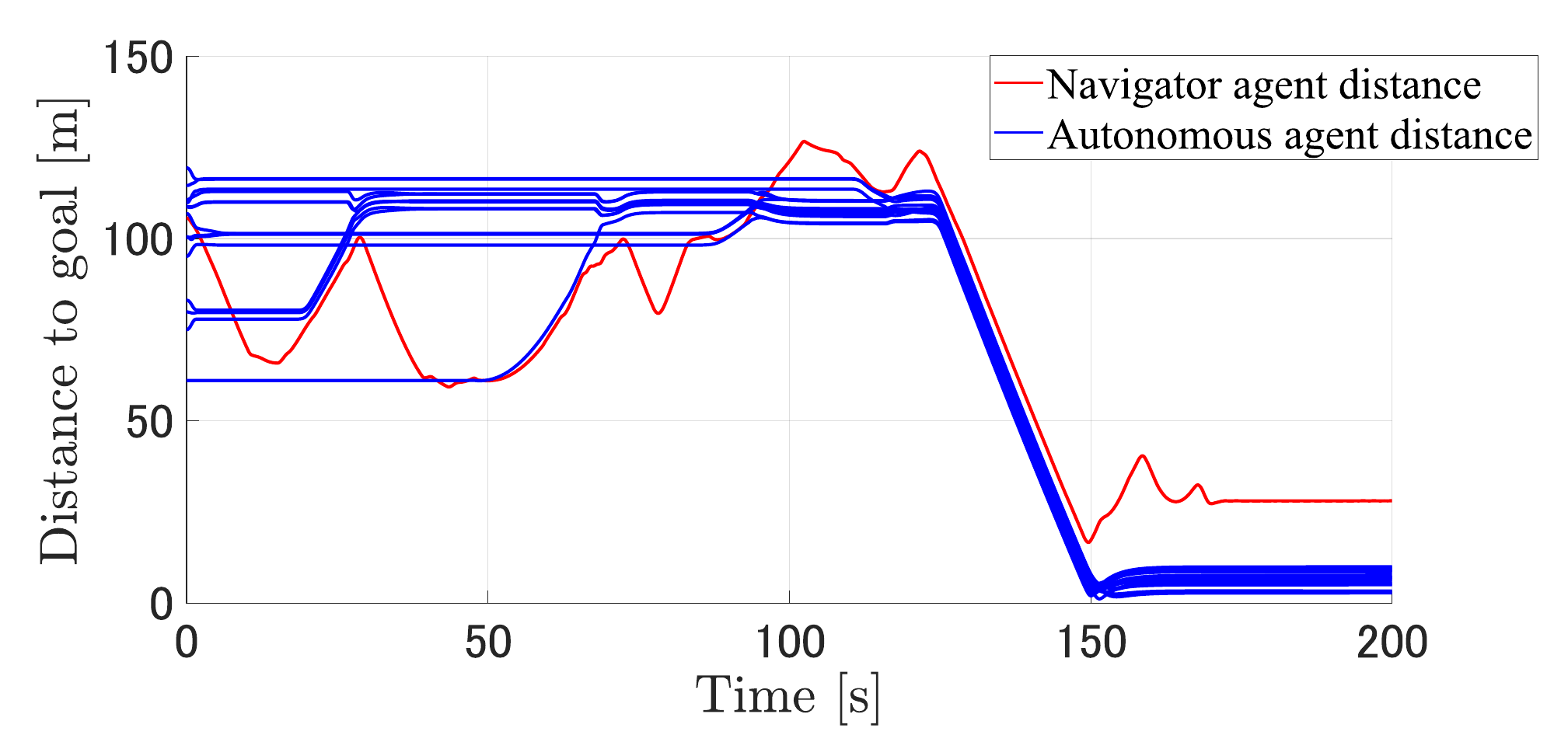}
    \caption{Distance to goal (sparse initial positions)}
    \label{fig:bunsan_dist}
\end{figure}

To verify the effectiveness of the proposed model, we conducted numerical simulations.
The simulation uses MATLAB 2025b, and the numerical solution employs a Euler method.
The simulation duration was set to $200$~[s], and the sampling period was set to $\Delta t = 0.06$~[s].
The simulation environment and initial states of each agent are shown in Fig.~\ref{fig:bunsan_initPos}.
Blue circles represent autonomous agents, red stars represent guided agents, and black crosses denote the goal location $\bm{p}_\mathrm{goal}$.
We consider a scenario with one navigator agent and $14$ autonomous agents, where the autonomous agents were initially dispersed.
The goal position $\bm{p}_{\mathrm{goal}}$ is $[0, -100, 40]^\mathrm{T}$~[m].
Fig.~\ref{fig:sim_result2_snap}, Fig.~\ref{fig:sim_result_sparse}, and Fig.~\ref{fig:bunsan_dist} show the guidance snapshot, guidance trajectory, and the evolution of each agent's distance to its target location, respectively.
As shown in Fig.~\ref{fig:sim_result_sparse}, we confirm that the navigator agent collects the separated clusters and guides the swarm toward the goal location.
Furthermore, as shown in Fig.~\ref{fig:bunsan_dist}, navigation is completed in approximately 150 [s] despite being distributed across multiple clusters.


\section{Conclusion}
\label{sec:conclusion}
This study proposed a prediction-based shepherding model that leverages motion constraints of autonomous agents, with the aim of application to drone-swarm guidance.
By combining clustering-based target selection with DWA-inspired prediction and motion selection, we developed a comprehensive guidance strategy.
Numerical simulations confirmed that, even when the autonomous agents are initially dispersed, the navigator agent can guide the swarm without leaving agents behind.
Future work will address the optimal design of the proposed evaluation function and its implementation on real drone swarms.

\section*{Acknowledgements}
This research was supported in part by grants-in-aid for JSPS KAKENHI Grant Number JP25K17629 and JST Moonshot Research and Development Program JPMJMS2032 (Innovation in Construction of Infrastructure with Cooperative AI and Multi Robots Adapting to Various Environments).

\bibliographystyle{tfnlm}
\bibliography{references}

\end{document}